# Natural Language Guided Visual Relationship Detection

Wentong Liao, Lin Shuai, Bodo Rosenhahn
Institute of Information Processing
Leibniz University Hanover
liao@tnt.uni-hannover.de

Michael Ying Yang
Scene Understanding Group
University of Twente
michael.yang@utwente.nl

## Abstract

*Reasoning about the relationships between object pairs in images is a crucial task for holistic scene understanding. Most of the existing works treat this task as a pure visual classification task: each type of relationship or phrase is classified as a relation category based on the extracted visual features. However, each kind of relationships has a wide variety of object combination and each pair of objects has diverse interactions. Obtaining sufficient training samples for all possible relationship categories is difficult and expensive. In this work, we propose a natural language guided framework to tackle this problem. We propose to use a generic bi-directional recurrent neural network to predict the semantic connection between the participating objects in the relationship from the aspect of natural language. The proposed simple method achieves the state-of-the-art on the Visual Relationship Detection (VRD) and Visual Genome datasets, especially when predicting unseen relationships (e.g., recall improved from 76.42% to 89.79% on VRD zero-shot testing set).*

## 1. Introduction

Scene understanding is one of the most primal topics in the computer vision and machine learning communities. It ranges from the pure vision tasks, such as object classification/detection [18, 31], semantic segmentation [23, 43], to the comprehensive visual-language tasks, e.g., image/region caption [16, 39], scene graph generation [40, 38, 21], and visual question-answering [3, 37]. Boosted by the impressive development of deep learning, the research of pure version tasks is becoming gradually mature. However, it is still challenging to let the machine understand the scene at a higher semantic level. Visual relationship detection is a promising intermediate task to bridge the gap between the vision and the visual-language tasks and has caught increasing attention [32, 24, 21].

Visual relationship detection targets on understanding the visually observable interactions between the detected objects in images. The relationships can be represented in a triplet form of ⟨*subject-predicate-object*⟩, e.g., ⟨*kid-on-skate board*⟩, as shown in Fig. 1. A natural approach for this task is to treat it as a classification problem: each kind of relationships/phrase is a relation category [32], as shown in Fig. 2. To train such reliable and robust model, sufficient training samples for each possible ⟨*subject-predicate-object*⟩ combination are essential. Consider the Visual Relationship Dataset (VRD) [24], with $N = 100$ object categories and $K = 70$ predicates, then there are $N^2K = 700k$ combinations in total. However, it contains only $38k$ relationships, which means that each combination has less than 1 sample on average. The previous classification-based works can only detect the most common relationships, e.g., [32] studied only 13 frequent relationships.

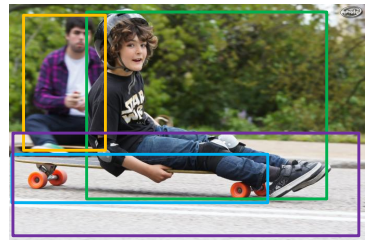

Figure 1: Visual relationships represent the interactions between observed objects. Each relationship has three elements: *subject*, *predicate* and *object*. Here is an example image from Visual Genome [17]. Our proposed method is able to effectively detect numerous kinds of different relationships from such image.

Another popular strategy is to detect the relationship predicates and the object categories independently. Although the number of categories decreases dramatically, the semantic relationship between the objects and the predicates are ignored. Consequently, the phrase which has the same predicate but different agents is considered as the same type of relationship. For instance, the "clock-on-

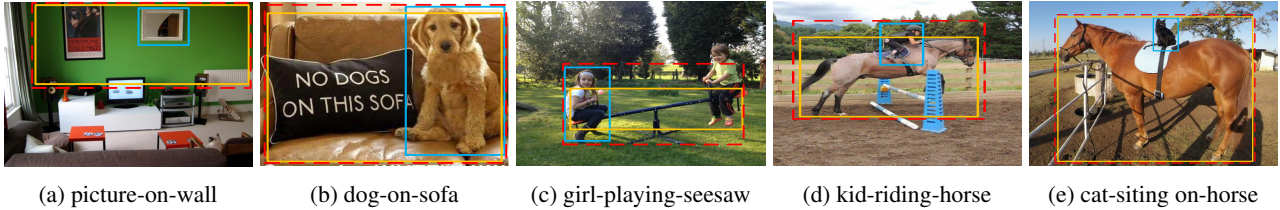

| (a) picture-on-wall | (b) dog-on-sofa | (c) girl-playing-seesaw | (d) kid-riding-horse | (e) cat-siting on-horse |

Figure 2: Examples of the wide variety of visual relationships, and its difference with the phrases. The solid bounding boxes indicate the individual objects and the dash red bounding boxes denote a phrase.

*wall"* (Fig. 2a) and *"dog-on-sofa"* (Fig. 2b) belong to the same predicate type *"on"*, but they describe different semantic scenes. On the other hand, the type of relationship between two objects is not only determined by their relative spatial information but also their categories. For example, the relative position of between the kid and the horse (Fig. 2d) is very similar as the ones between the cat and the horse (Fig. 2e), but it is preferred to describe the relationship *"cat-sitting on-horse"* rather than *"cat-riding-horse"* in the natural language setting. It's also very rare to say *"person-sitting on-horse"*.

Another important observation is that the relationships between the observed objects are naturally based on our language knowledge. For example, we would like to say the kind "sitting on" or "playing" the seesaw but not "riding" (Fig. 2c), even though it has the very similar pose as that the kind "riding" the horse in Fig. 2d. On the other hand, similar categories have a similar semantic connection, for example, *"person-ride-horse"* and *"person-ride-elephant"*, because "horse" and "elephant" belong to the same category (animal). It is an important cue for inferring the infrequent relationships from the frequent instances. Fortunately, this semantic connection has been well researched in the language model [26, 27]: an object class is closed to another one if they belong to the same object category and far from the one belonging to a different category in the word-encoded space. The vivid example given in [26] *king-man=queen-woman* reveals that the inherent semantic connection between "king" and "man" is the same as "queen" and "woman". Here, "king" and "queen" belong to the same category (ruler) while "man" and "woman" in the same category (person). Therefore, we resort the powerful semantic connection in the language to handle the challenging problems in the task of visual relationship detection.

In this work, we propose a new framework for visual relationship detection in large-scale datasets. The visual relationship detection task is roughly divided into two subtasks. The first task is to recognize and localize objects that are present in a given image. It provides the visual cues of *"what"* and *"where"* are the objects. The second task is to reason about the interaction between an arbitrary pair of the observed objects. It understands *"how"* they connect with each other in a semantic context. We show that our model is able to scale and detect thousands of relationship types by leveraging the semantic dependencies from language knowledge, especially to infer the infrequent relationships from the frequent ones.

The major **contributions** of this work are as follows:

1. We propose to use a generic bi-directional recurrent neural network (RNN) [33, 25] to predict the semantic connection, *e.g.*, predicate, between the participating objects in the relationship from the aspect of natural language knowledge.

2. The natural language knowledge can be learned from any public accessible raw text, *e.g.*, the image captions of a dataset.

3. The visual features of the union boxes of the two participating objects in the relationships are not required in our method. State-of-the-art methods [24, 20, 21, 41, 7] rely on such features. Furthermore, our method is able to infer infrequent relationships from the frequent relationship instances.

4. Our model is competitive with the state-of-the-art in visual relationship detection in the benchmark datasets of Visual Relationship Dataset [24] and Visual Genome [17], especially when predicting unseen relationships (*e.g.*, recall improved from 76.42% to 89.79% on VRD zero-shot testing set). The code will be made publicly available.

## 2. Related Work

As an intermediate task connecting vision and vision-language tasks, many works have attempted to explore the use of visual relationship for facilitating specific high-level tasks, such as image caption [5, 9], scene graph generation [40, 38], image retrieval [30], visual question and answering (VQA) [3, 2, 37], *etc*. Compared to these works which treat the visual relationship as an efficient tool for their specific tasks, our work dedicates to provide a robust framework for generic visual relationship detection.

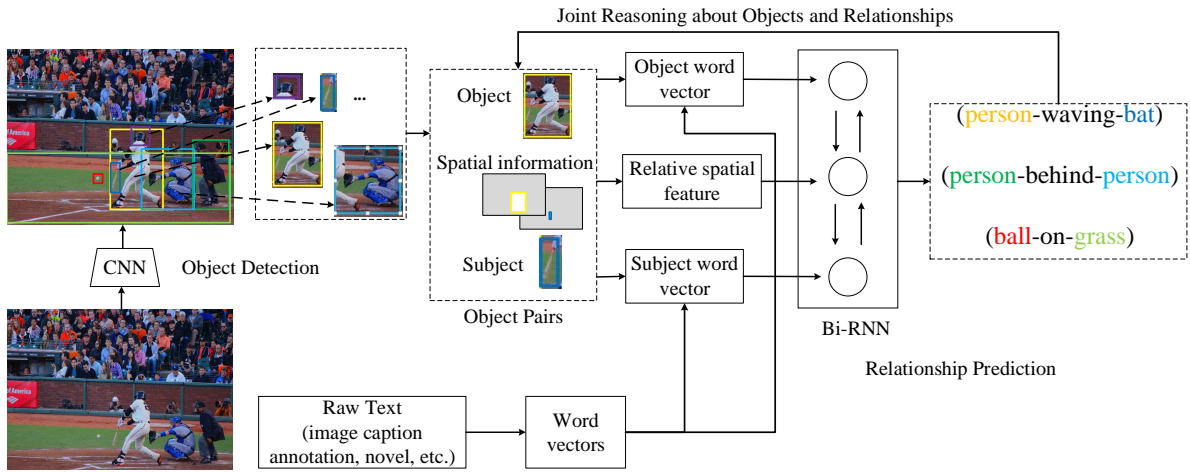

Figure 3: The proposed framework for visual relationship detection. First, Faster RCNN is utilized to localize objects and provide the classification probability of each detected object in the given image. Then, the possible meaningful object pairs are selected as the candidate relationships. Each object converted into the corresponding word vector.

Visual relationship detection is not a new concept in literature. [10, 12] attempted to learn four spatial relationships: "above", "below", "inside", and "around". [34, 40] detected the physical support relations between adjacent objects: support from "behind" and "below". In [32, 8], each possible combination of visual relationship is treated as a distinct visual phrase class, and the visual relationship detection is transformed to a classification task. Such methods suffer the long trail problem and can only detect a handful of the frequent visual relationships. Besides, all above works used the handcraft features.

In recent years, deep learning has shown its great power in learning visual features [18, 35, 14, 36, 23]. The most recent works [24, 38, 20, 21, 7, 42, 28, 29] use deep learning to extract representative visual features for visual relationship detection. In [38], the visual relationships are treated as the directed edges to connect two object nodes in the scene graph. The relationships are inferred along the processing of constructing the scene graph in an iterative way. [20, 21] focused on extracting more representative visual features for visual relationship detection, object detection, and image caption [21]. [7, 42] reasoned about the visual relationships based on the probabilistic output of object detection. [42] attempted to project the observed objects into relation space and then predict the relationship between them with a learned relation translation vector. [7] proposed a particular form of RNN (DR-Net) to exploit the statistical relations between the relationship predicate and the object categories, and then refine the estimates of posterior probabilities iteratively. It achieves substantial improvement over the existing works. However, most of the existing works [20, 21, 7] require additional union bounding boxes which cover the object and subject together to learn the visual features for relationship prediction. Besides, their works are mainly designed based on visual aspect. In this paper, we analyze the visual relationships from the language aspect. The most related works are [24, 41, 29], which proposed to use linguistic cues for visual relationship detection. [24] attempted to find a relation projection function to transform the word vectors [26] of the participating objects into the relation vector space for relationship prediction. [41] exploited the role of both visual and linguistic representations and used internal and external linguistic knowledge to regularize the network's learning process. [29] proposed a framework for extracting visual cues from a given image and linguistic cues from the corresponding image caption comprehensively. In particular, for visual relationship detection in the VRD dataset, 6 Canonical Correlation Analysis [11] models are trained. Different from their works, our method uses a modified BRNN to leverage the natural language knowledge, which is much simpler and outperforms [11] regarding visual relationships prediction.

## 3. Visual Relationship Prediction

The general expression of visual relationships is ⟨*subject-predicate-object*⟩. The component "predicate" can be an action (*e.g.* "wear"), or relative position (*e.g.* "behind"), *etc*. For convenience, we adopt the widely used convention [32, 24] to characterize each visual relationship in the triplet form as ⟨*s-p-o*⟩, such as ⟨*person-wave-bat*⟩. Here, $s$ and $o$ indicate the subject and object category respectively, while $r$ denotes the relationship predicate. Con-cretely, the task is to detect and localize all objects presented in an image and predict all possible visual relationships between any two of the observed objects. Note that, "no relation" is also a kind of visual relationship between two objects in this work. For instance, in Fig. 2e, there is no explicit visual relationship between the "cat" and the "tree". An overview of our proposed framework is shown in Fig. 3. It comprises multiple steps, as described as follows.

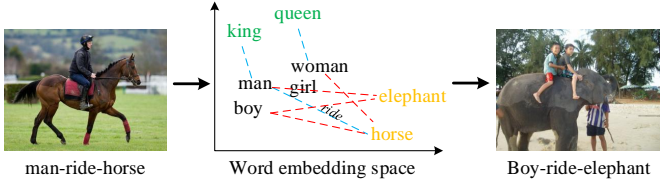

Figure 4: Example of inferring infrequent relationships (the left image) from frequent instances (the right image) guided by natural language knowledge. In the middle image, the blue dashed lines denote the distances between words. We assume this distance as the inherent semantic connection in natural language knowledge. The infrequent relationships, which is connected with red dashed lines, can be inferred from the frequent relationships.

### 3.1. Object detection

Before reasoning about the visual relationships, objects present in the given images must be localized as a set of candidate components of the relationships. In this work, the Faster RCNN [31] is utilized for this task because of its high accuracy and efficiency. Each detected object comes with a bounding box to indicate its spatial information, and the object classification probabilities $\mathbf{p_o} = \{p_i\}_{i=1,\cdots,N+1}$, where $p_i$ is the predicted probability that the object belongs to object category $i$, $N$ is the total number of object categories and $N+1$ indicates the background object. And the location of each detected object is denoted as $(X, Y, W, H)$, where $(X, Y)$ is the normalized coordinate of the up-left corner point of the bounding box on the image plane, and $(W, H)$ is the normalized *width* and *height* of the bounding box.

### 3.2. Natural language guided relationship recognition

The word vectors embed the semantic context between different words in a semantic space [26, 27]. The words which have similar semantic meaning are close to each other in the space, for example as shown in the middle image of Fig. 4. On the other side, the distances between the words in a semantic group and the words in different semantic groups could be similar. Even though the distance between different words in the embedded word space is cal-culated as cosine distance [26], we assume that it is inherent semantic relationships connecting the two words rather than a mathematics distance in the embedding space. For example, the semantic connection between "person" and "horse" is normally "ride". "horse" and "elephant" are in the same semantic group (*animal*). Therefore, "ride" is very likely the semantic connection between "horse" and "elephant". This semantic property is important to lean the infrequent relationship (*e.g.*, "person ride elephant, camel, tiger, *etc.*") from the very normal relationship ("person ride horse") in the real world. Fig. 4 illustrates a brief process of this inference.

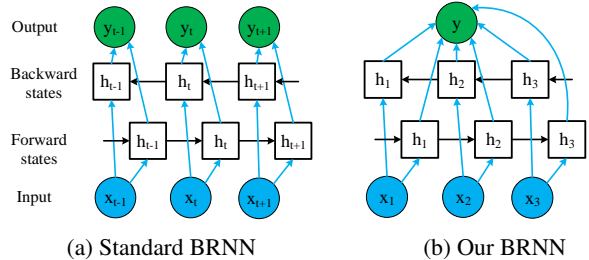

(a) Standard BRNN          (b) Our BRNN

Figure 5: The standard BRNN model [33] (a), and (b) our BRNN model used for predicate prediction. Our BRNN has three inputs in sequence(*subject*, *spatial information* and *object*) and one output (predicate prediction).

Bi-directional RNNs (BRNNs) [33] have achieved great successes for natural language processing tasks [15, 4, 6, 13]. The standard BRNN structure is shown in Fig. 5a. The vector $x_t$ is the input of a sequence at time point $t$ and $y_t$ is the corresponding output, while $h_t$ is the hidden layer. A BRNN computes the hidden states twice: a *forward* sequence $\overrightarrow{h}$ and a *backward* sequence $\overleftarrow{h}$. Each component can be expressed as follows:

$$\overrightarrow{h}_t = \mathcal{H}(W_{x\overrightarrow{h}}x_t + W_{\overrightarrow{h}\overrightarrow{h}}\overrightarrow{h}_{t-1} + b_{\overrightarrow{h}}), \quad (1)$$

$$\overleftarrow{h}_t = \mathcal{H}(W_{x\overleftarrow{h}}x_t + W_{\overleftarrow{h}\overleftarrow{h}}\overleftarrow{h}_{t+1} + b_{\overleftarrow{h}}), \quad (2)$$

$$y_t = W_{\overrightarrow{h}y}\overrightarrow{h}_t + W_{\overleftarrow{h}y}\overleftarrow{h}_t + b_y. \quad (3)$$

where $W_{x\overrightarrow{h}}$ denotes the input-hidden weight matrix in forward direction. $b_{\overrightarrow{h}}$ denotes the bias vector of the hidden layer in forward direction. $\mathcal{H}$ is the activation function of the hidden layers. We use the ReLU function [19] in this work. The output sequence $\mathbf{y}$ is computed by iterating the considering both of the forward and backward input sequence $\mathbf{x}$. This process plays an important role in visual relationship detection. Since in a relationship expression $\langle subject\text{-}predicate\text{-}object\rangle$, the order of the two objects is decisive for the final prediction. *E.g.*, $\langle person\text{-}ride\text{-}horse\rangle$ is completely different from $\langle horse\text{-}ride\text{-}person\rangle$. A BRNN is able to learn such difference caused by the input sequence.

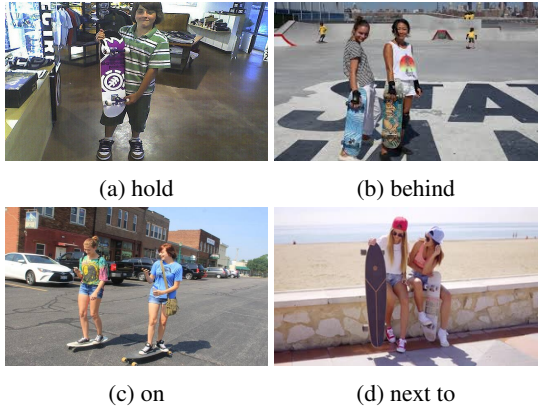

(a) hold     (b) behind

(c) on     (d) next to

Figure 6: The relative position of the two objects is crucial for the relationship prediction.

This is the main reason why we used BRNN to learn the linguistic cues between the objects categories.

Besides the object categories, the relative position of the participating objects is crucial for predicate prediction, as illustrated in Fig. 6. Even though the object categories are the same in all instances, the predicate between the *"person"* and the *"skateboard"* is different in each image. Therefore, we modify the standard BRNN, as shown in Fig. 5b. The input terms $x_1$ and $x_3$ are the word vectors (300-dimension of each) of the participating objects, respectively. In this work, the *Glove* algorithm [27] is used to learn the word vectors because of its high efficiency and robust performance. $x_2$ is the spatial information of the two objects. As introduced in Sec. 3.1, each detected object is localized with $(X, Y, W, H)$. We concatenate the two locations and pad zeros to construct the 300-dimension input $x_2$. Compared to the CNN used in [7] to extract the spatial features, the simple spatial information $x_2$ gives similar performance in our framework yet with much less parameters. From this aspect, our method saves part of the computation to train the CNN for spatial feature learning. Therefore, Eq. (3) are redefined in our framework as:

$$\begin{aligned} y = & (W_{\overrightarrow{h_1}y}\overrightarrow{h_1} + W_{\overrightarrow{h_2}y}\overrightarrow{h_2} + W_{\overrightarrow{h_3}y}\overrightarrow{h_3}) \\ & + (W_{\overleftarrow{h_1}y}\overleftarrow{h_1} + W_{\overleftarrow{h_2}y}\overleftarrow{h_2} + W_{\overleftarrow{h_3}y}\overleftarrow{h_3}) \\ & + b_y. \end{aligned} \quad (4)$$

The first term is for information passed from the *forward* sequence and the second term is for *backward* sequence. Each notation in the equation has the similar meaning as in the standard BRNN. The output $y$ is $K+1$-dimension, where $K$ is the number of predicate types. The $(K+1)th$ predicate type is *"no relationship"*. We use the softmax function on $y$ to compute the normalized probabilistic prediction of the predicate. As described in Eq. (1), $\overrightarrow{h_0}$ is required for computing $\overrightarrow{h_1}$. We set $\overrightarrow{h_0}$ as zero-vector, and so as for $\overleftarrow{h_4}$ when $\overleftarrow{h_3}$ is computed.

The raw text can be any public accessible text data, such as the ground truth region caption of the Visual Genome dataset [17]. This step is performed at the beginning of the training phase. In Sec. 4, we will study the influence of the word vectors learned from different raw text.

### 3.3. Joint recognition

At the test time, the categories of detected objects and the predicate types are jointly recognized. The joint probability of relationships and object detection can be written as:

$$p(O_s, r, O_o) = p(O_s)p(r|O_s, O_o)p(O_o). \quad (5)$$

where $p(O_s)$ and $p(O_o)$ are the probabilities of predicted categories of the *subject* and *object* in the relationship, respectively. They are given by the Faster RCNN [31]. $p(r|O_s, O_o)$ is the probabilities of predicted predicate given by the BRNN. On each test image, we find the optimal prediction using:

$$\langle O_s^*, r^*, O_o^* \rangle = \arg\max_{O_s, r, O_o} p(O_s, r, O_o) \quad (6)$$

## 4. Experiments

We evaluated our model on two datasets: (1) **VRD** [24]: the dataset contains $5,000$ images with 100 object categories and 70 predicates. There are $37,993$ visual relationship instances that belong to $6,672$ relationship types. We follow the train/test split in [24], i.e. $4,000$ images for training and $1,000$ images for testing. VRD has a zero-shot testing set that contains relationships that never occur in the training data. We evaluate on the zero-shot sets to demonstrate the generalization improvements brought by natural language knowledge. (2) **Visual Genome (VG)** [17]: the dataset contains 108K images and 998K relationships that belong to $74,361$ relationship types. We follow [21] to pre-process the dataset: the top-150 frequent object categories and top 50 predicate categories. The ground truth annotation whose shorter edges of the bounding boxes are smaller than 16 pixels are removed. Because the object detection network downscales the input image into one 32nd of the original size. Those objects will disappear in the final feature maps. After the preprocessing step, $95,998$ images are left. $25,000$ images are randomly selected as the testing set and the remaining $70,998$ images are used for training [21].

### 4.1. Experiment settings

**Model training details.** In all experiments, our model was trained using TensorFlow [1]. The Faster RCNN model is initialized on the ImageNet pretrained VGG-16 network [35] and is optimized using SGD. The BRNN model has two hidden layers, and each of them has 128 hidden states. Its parameters are initialized randomly. The word vectors (word2vec) are learned by *Glove* [27] from the ground truth region caption in VG [17].

| Dataset | Comparison | Predicate Detection | | Phrase Detection | | Relationship Detection | |
|---|---|---|---|---|---|---|---|
| | | Rec@50 | Rec@100 | Rec@50 | Rec@100 | Rec@50 | Rec@100 |
| VG | LP [24] | 26.67 | 33.32 | 10.11 | 12.64 | 0.08 | 0.14 |
| | SG [38] | 58.17 | 62.74 | 18.77 | 20.23 | 7.09 | 9.91 |
| | MSDN [21] | 67.03 | 71.01 | 24.34 | 26.50 | 10.72 | 14.22 |
| | DR-Net [7] | **88.26** | 91.26 | 23.95 | 27.57 | 20.79 | **23.76** |
| | Ours | 85.21 | 91.56 | 43.51 | **46.09** | **21.49** | 23.51 |
| | Ours+(COCO [22]) | 83.52 | **92.04** | **43.72** | 46.01 | 21.03 | 23.51 |
| VRD | LP [24] | 47.87 | 47.87 | 17.03 | 16.17 | 14.70 | 13.86 |
| | CCA [29] | - | - | 16.89 | 20.70 | 15.08 | 18.37 |
| | DR-Net [7] | 80.78 | 81.90 | 19.93 | 23.45 | 17.73 | 20.88 |
| | LK [41] | **85.64** | **94.65** | 26.32 | 29.43 | **22.68** | **31.89** |
| | Ours | 84.92 | 92.65 | **42.29** | **47.92** | 20.81 | 22.22 |
| | Ours+(COCO [22]) | 82.49 | 90.85 | 41.36 | 45.83 | 20.92 | 23.21 |

Table 1: Experimental results of different methods in the VRD [24] and VG [17]. We compare our method with the existing works on the three tasks discussed in Sec. 4.1.

**Performance metrics.** We follow the metrics for visual relationship detection [38]: the *Top-K* recall, which is denoted as *Rec@K*, as the main performance metric, which is the fraction of ground truth instance which fall in the *Top-K* predictions. We will report mainly the *Rec@50* and *Rec@100* in the experiments.

**Task settings.** Visual relationship detection involves localizing and classifying both the objects and predicting the predicate. We evaluate our model on three convention tasks [38]: (1) **Predicate detection**: In this task, the ground truth locations and labels of both *subject* and *object* are given. This task aims at measuring the accuracy of predicate recognition without the effect of the object detection algorithms. (2) **Phrase detection**: The input is an image and the ground truth locations of individual objects. The output is a set of relationships ⟨*subject-predicate-object*⟩ and the union bounding boxes which covers the whole triplet. When all the three elements of a relationship are correctly predicted and the IoU between the predicted union boxes and their ground truth is above 0.5, this prediction is considered as correct. This task evaluates the model ability of object classification and predicate prediction. (3) **Relationship detection**: Given an image, a set of relationships ⟨*subject-predicate-object*⟩ are predicted. The IoU between the predicted locations and the ground truth boxes of both *subject* and *object* are at least 0.5 simultaneously. This task evaluates the model for both object and predicate detection.

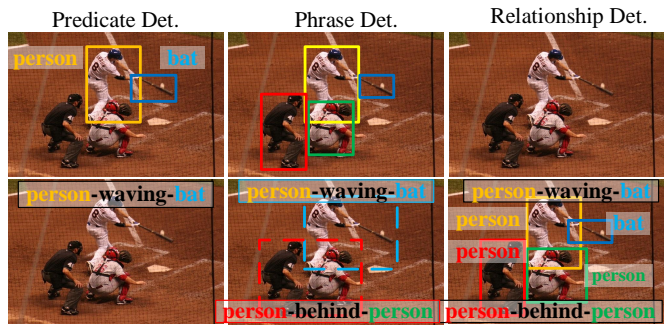

Figure 7: Illustration of different task settings. The first row depicts the inputs for different tasks and the second row is the corresponding outputs. The solid bounding boxes localize individual objects while the dashed bounding boxes localize the locations of phrases.

An illustration for different task settings is shown in Fig. 7.

### 4.2. Comparative results

Our method is compared with the following methods extensively under the three task settings: (1) **LP** [24]: the representative work which uses language prior for predicate estimation. (2) **SG** [38]: the representative work which detects visual relationships on the VG dataset. (3) **MSDN** [21]: the most recent work which focuses on refining visual features for relationship detection and image caption. (4) **DR-Net** [7]: the current state-of-the-art which gives the best performance in all three tasks in both VRD and VG datasets. (5) **CCA** [29]: the most recent work which also uses the linguistic cues associating with the visual cues for visual relation detection. (6) **LK** [41]: another recent work which exploits the role of both visual and linguistic representations. (7) Furthermore, we train the word vector using the image caption from COCO dataset [22]. This comparison is to evaluate the generalization ability and the robustness of our method when the sources of language knowledge are different.

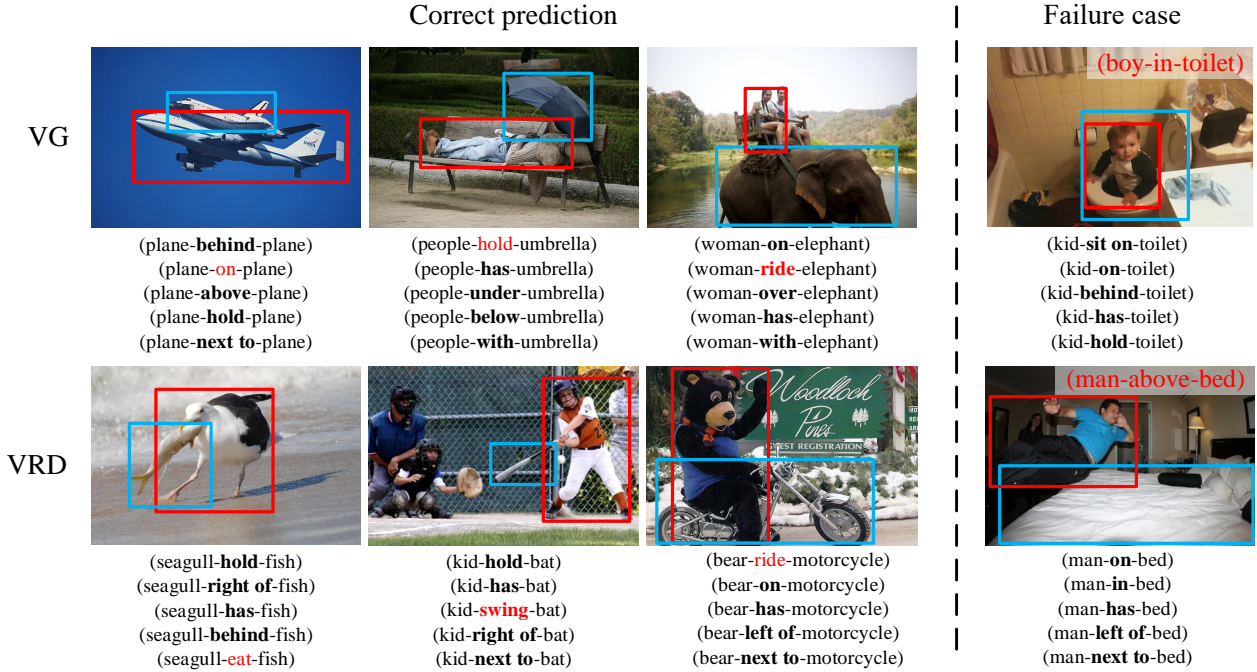

Figure 8: Qualitative examples of visual relationship detection on VG [17] (the first row) and VRD [24] (the second row) respectively. The first three columns illustrate the correctly recognized relationships in the images, while the last column is the failure example (the ground truth relationship phrase is denoted in red in the image). The red bounding boxes denote the subjects while the cyan boxes denote the objects. The relationships under the images are the $Top-5$ most probable relationships predicted by our method, in which the red denotes the ground truth.

| predicate | [38] | Ours | predicate | [38] | Ours |
|---|---|---|---|---|---|
| on | **99.71** | 99.39 | under | 56.93 | **83.44** |
| has | 96.47 | **98.47** | sitting on | 57.01 | **91.07** |
| in | 88.77 | **93.87** | standing on | 61.90 | **78.06** |
| of | 96.18 | **97.80** | in front of | 64.63 | **75.67** |
| wearing | 98.01 | **99.59** | attached to | 27.43 | **70.00** |
| near | 95.14 | **99.57** | at | 70.41 | **86.33** |
| with | 88.00 | **93.69** | hanging from | 0 | **67.50** |
| above | 70.94 | **86.33** | over | 0.69 | **56.00** |
| holding | 82.80 | **96.18** | for | 11.21 | **57.22** |
| behind | 84.12 | **93.30** | riding | 91.18 | **95.08** |

Table 2: The per-type predicate classification accuracy with metric Rec@5. These predicate types are the $Top-20$ most frequent cases in the dataset (sorted in descending order in the table).

Table 1 shows the results. Our method outperforms all the existing works in the three task settings in both datasets, except [7, 41] who perform slightly better than ours on some tasks.

From Table 1, one can observe that, even though the improvements for predicate detection from SG [38] to DR-Net [7] are significant (28.52% for *Rec@100* on VG), the improvements for phrase detection are much less (7.34% for *Rec@100* on VG). Because [7, 38] only use visual features for visual relationship detection. The drawback of pure vision model for visual relationship detection is that it cannot analyze which two object categories are more likely to have semantic connection. On the other side, our method achieves substantial improvements on the phrase detection task (18.52% for *Rec@100* on VG). These results show that our method can effectively pair the objects which have essential relationships and precisely predict their predicate in the images. LP [24] also used extracted word vectors for visual relationship detection. However, the linear projection function in their model, which transforms the objects categories into the relationship vector space, is inadequate for predicting numerous kinds of relationships. In contrast, our BRNN model includes multiple nonlin-ear activation function which can learn more representative features than LP [24]. Our method also outperforms CCA [29], which also used linguistic cues for visual relationship detection. While the performance of our method is inferior to LK [41] on the tasks of predicate detection (2% for *Rec@100*) and relationship detection (9.67% for *Rec@100*), our approach performs much better on the task of phrase detection (18.49% for *Rec@100*).

|  | Predicate Detection | | Phrase Detection | | Relationship Detection | |
| --- | --- | --- | --- | --- | --- | --- |
|  | Rec@50 | Rec@100 | Rec@50 | Rec@100 | Rec@50 | Rec@100 |
| LP [24] | 8.45 | 8.45 | 3.75 | 3.36 | 3.52 | 3.13 |
| CCA [29] | - | - | 10.86 | 15.23 | 9.67 | 13.43 |
| LK [41] | 56.81 | 76.42 | 13.41 | 17.88 | 12.29 | 16.37 |
| Ours | **80.25** | **89.79** | **42.10** | **42.51** | **19.71** | **21.97** |

Table 3: Experimental results for zero-shot visual relationship detection on the VRD dataset [24].

Fig. 8 shows some qualitative examples of visual relationship detection in the two datasets. From the first three columns we can see that the $Top-5$ most probable predicted predicates between the objects are highly close to the ground truth, *e.g.*, *on* is very close to *ride* from the spatial aspect. The rare relationship of ⟨bear-ride-motorcycle⟩ is successfully predicted with the highest probability, which shows that our model can learn very rare relationships from the normal relationships guided by natural language knowledge. The last column gives a failure example of each dataset. ⟨kid-in-toilet⟩ and ⟨man-above-bed⟩ are both abnormal scenes in the real world. The natural language knowledge extracted by our model guides the prediction towards more probable results. Our current model fails to detect abnormal interactions in the natural scenes.

Furthermore, we compared the results of our method by using different raw text to learn word vectors. Besides VG, we also trained the word vector using the image caption from COCO dataset [22], as shown in Table 1 (Ours+(COCO)). We observe that our model performs robustly, sometimes better than using the raw text of the VG dataset. The main reason is that COCO dataset provides richer manual image cation annotation (5 captions per image). Therefore, the learned word vectors contain more information about the semantic connection. These experiments show that using natural language knowledge for visual relationship detection is an effective and robust scheme. There are many large-scale image datasets for object detection. But most of them don't have the annotation for visual relationship detection. It could be an effective solution to generate high-quality annotation of visual relationships using natural language knowledge automatically.

Table 2 shows the performance on predicting per-type predicate of SG [38] and our method. These results are calculated in the task of predicate detection. Our method reports much better results than SG [38] on each type predicate classification, in particular, 67.50% improvement on the type *hanging from*. This table shows that our model performs well in predicting frequent predicates.

### 4.3. Zero-shot learning

In the VRD dataset [24], there are $1,877$ relationships in the test set that have never occurred in the training set. Our trained model is utilized to detect these unseen relationships to evaluate its ability of inferring infrequent relationships based on frequent relationships that have ever seen, namely *zero-shot learning*. Table 3 shows the results from different works. Our method outperforms LP [24], CCA [29] and LK [41] by large margin, while only decreasing slightly compared with the results shown in Table 1. This table shows that our model has good generalization ability: it can detect thousands of relationships, even the instances that have never been seen.

## 5. Conclusion

This paper presents a natural language knowledge guided method for detecting visual relationships in images. The semantic connection between the object categories and predicate are embedded in the word vector learned by natural language processing. We designed a BRNN model to predict the predicate between two observed objects based on this natural language knowledge and their spatial information. In particular, our method is able to infer infrequent relationships from the frequent relationship instances, which is important to deal with the long tail problem. Experiments on the Visual Genome and Visual Relationship Datasets show substantial improvements compared with most existing works for visual relationship detection in terms of accuracy and generalization ability. In the zero shot learning task, the proposed method shows the potential for detecting thousands of relationships. In the future work, we will extend our current model to an end-to-end framework which can learn better representative visual features from image and language features from raw text simultaneously. Another interesting direction is learning multiple tasks with a single network, such as object detection, visual relationship detection and image captioning.

## 6. Supplementary Note

The supplementary material is not necessary to understand the paper. This supplementary note discusses the following points:

- Further details on our training procedure.
- Detailed experimental results.
- Additional images with qualitative results.

### 6.1. Further Details on our Training Procedure

| Predicate | Results | Predicate | Results |
|---|---|---|---|
| on | 99.39 | eating | 87.80 |
| wearing | 99.59 | belonging to | 92.88 |
| has | 98.47 | parked on | 86.36 |
| of | 97.80 | hanging from | 67.50 |
| in | 93.87 | to | 30.45 |
| near | 99.57 | between | 32.47 |
| behind | 93.30 | covering | 70.24 |
| holding | 96.18 | playing | 48.15 |
| with | 93.69 | covered in | 88.46 |
| above | 86.33 | along | 56.50 |
| sitting on | 91.07 | on back of | 63.85 |
| riding | 95.08 | lying on | 35.13 |
| under | 83.44 | part of | 31.60 |
| in front of | 75.67 | using | 54.00 |
| and | 34.50 | walking in | 46.80 |
| standing on | 78.06 | mounted on | 50.00 |
| at | 86.33 | from | 18.75 |
| carrying | 83.61 | growing on | 14.51 |
| attached to | 70.00 | painted on | 38.89 |
| walking on | 93.83 | made of | 60.00 |
| over | 56.00 | flying in | 0 |
| looking at | 68.18 | says | 0 |
| for | 57.22 | across | 0 |
| watching | 68.15 | against | 0 |
| laying on | 65.10 | | |

Table 4: The per-type predicate classification accuracy with metric *Rec@5*. These predicate types are all cases in the Visual Genome (**VG**) dataset [17] (sorted in descending order in the table, as shown in Fig. 9).

We will now discuss the training procedure of our framework. We use the *TensorFlow* framework to implement our framework. Our training procedure consists of two parts: training the Faster RCNN [31] for object detection and training the BRNN for predicate prediction.

**Faster RCNN Training.** The Faster RCNN network (without the final classification layer) is initialized with the parameters pre-trained on ImageNet, and the parameters of the final classification layer are randomly initialized. Then, it is fine-tuned on the **VG** [17] and **VRD** [24], respectively. The fine-tuned details are as fellows. A mini-batch has one image. After the proposals are generated by RPN layers, we use 0.7 as the NMS threshold for object proposals and keep at most 2000 boxes after NMS. Then 256 object proposals are sampled: 128 for both of positive and negative samples (if the positive samples are not enough, more negative samples are sampled to complement the total number to 256). We use the *SGD* with gradient clipping provided by *TensorFlow* to train the network with the base learning rate 0.01. In testing, we set the NMS threshold to 0.4 for object proposals.

**BRNN Training.** The BRNN network is randomly initialized. The RNN unit has two hidden layers and each of them has 128 hidden states. All the parameters of BRNN are randomly initialized. We use the *SGD* with gradient clipping provided by *TensorFlow* to train the network with the base learning rate 0.01. The mini-batch size is 128.

### 6.2. Detailed Experimental Results

The detailed results of per-type predicate prediction on **VG** dataset are provided in Table 4. We merge the predicate "wears" with "wearing". Consequently, there are 49 predicate types. The occurrence number of each predicate type is shown in Fig. 9.

### 6.3. Additional Qualitative Results

Here we provide some additional qualitative results for the **VG** [17] and **VRD** [24] datasets. Fig. 10 shows some representative results for relationship detection on the **VG** [17] dataset: input is an image, and output are the detected objects and their relationships. Most of the miss detections are in fact due to the performance of object detector. Fig. 11 shows some representative results for *zero-shot learning* on the **VRD** dataset, which attempts to detect the relationships that have never occurred in the training set.

### Acknowledgments

The work is funded by DFG (German Research Foundation) YA 351/2-1 and RO 4804/2-1. The authors gratefully acknowledge the support. We thank NVIDIA for donating GPUs.

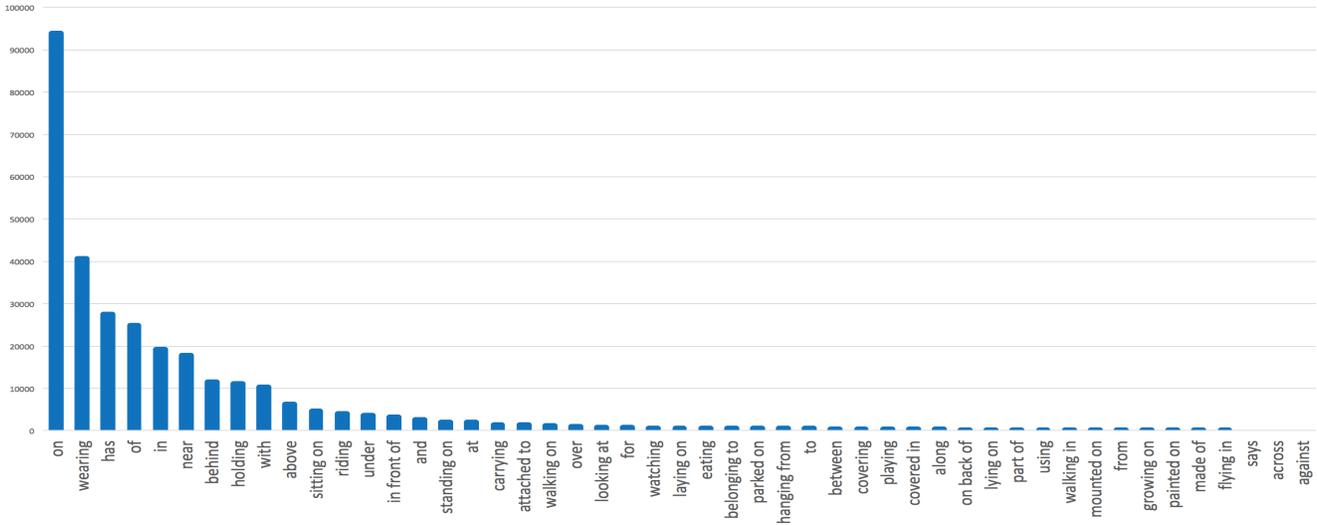

Figure 9: Statistic of each predicate type in the **VG** dataset. The $y$-axis denotes the occurrence number.

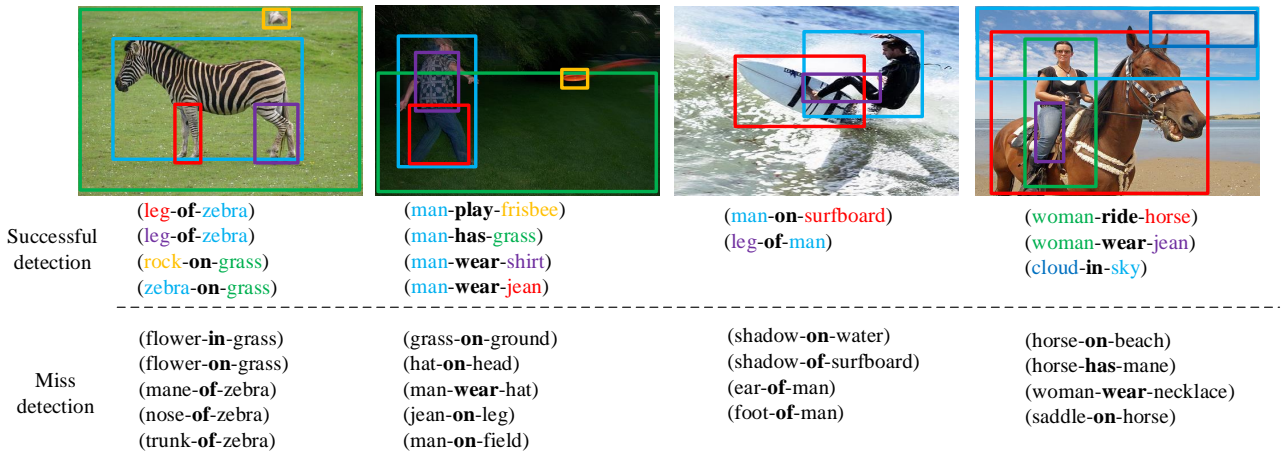

Figure 10: Qualitative results of *relationship prediction* on **VG** dataset [17]. The first row shows the example images and the objects detected by Faster RCNN [31]. The second row are the relationships which are correctly detected by our method. The third row are the relationships which are annotated in the ground truth but not detected by our method.

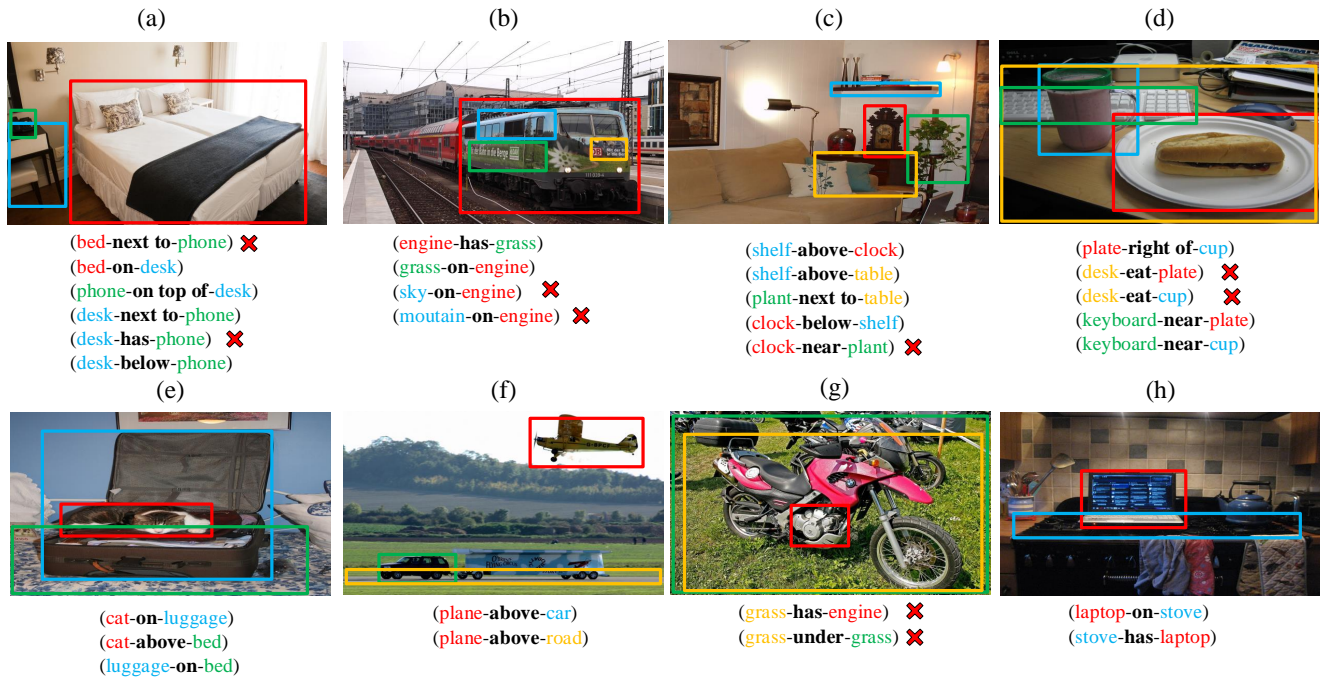

Figure 11: Qualitative results of *zero-shot learning* on **VRD** dataset [24]. The task setting is *phrase detection*. The first row shows the images and the objects annotation. The second row lists the ground truth zero-shot relationship instances. The red crosses denote that the relationships are missed by our method in the *Rec@50*, while the remains are correctly predicted.